\documentclass[conference]{IEEEtran}
\usepackage{times}

\usepackage[numbers]{natbib}
\usepackage{multicol}
\usepackage[bookmarks=true]{hyperref}
\usepackage{booktabs}
\usepackage{amssymb}
\usepackage{pifont}
\usepackage{multicol}
\usepackage{multirow}
\usepackage{color}
\usepackage[dvipsnames]{xcolor}
\usepackage{graphicx}
\newcommand{\yes}{\large \color{OliveGreen}\checkmark}
\newcommand{\no}{\color{BrickRed} \scalebox{1}{\usym{2613}}}
\usepackage[normalem]{ulem}
\usepackage{subcaption}
\usepackage{utfsym}
\usepackage{fontawesome}
\usepackage{marvosym}
\usepackage{adjustbox}
\definecolor{cerulean}{rgb}{0.0, 0.48, 0.65}
\makeatletter
\def\blfootnote{\xdef\@thefnmark{}\@footnotetext}
\makeatother

\newcommand{\methodname}{TeleMoMa}

\begin{document}

\title{\methodname{}: A Modular and Versatile Teleoperation System for Mobile Manipulation}

\author{
Shivin Dass$^{1}$,
Wensi Ai$^{2}$,
Yuqian Jiang$^{1}$,
Samik Singh$^{1}$,
Jiaheng Hu$^{1}$, \\
Ruohan Zhang$^{2}$,
Peter Stone$^{1, 3}$,
Ben Abbatematteo$^{1}$,
Roberto Mart{\'i}n-Mart{\'i}n$^{1}$ 
\\
\\
$^{1}$The University of Texas at Austin \hspace{0.2in} $^{2}$Stanford University \hspace{0.2in} $^{3}$Sony AI \hspace{0.2in}\\
\vspace{-2em}
}

\maketitle

\begin{abstract}
A critical bottleneck limiting imitation learning in robotics is the lack of data. This problem is more severe in mobile manipulation, where collecting demonstrations is harder than in stationary manipulation due to the lack of available and easy-to-use teleoperation interfaces. In this work, we demonstrate \methodname{}, a general and modular interface for whole-body teleoperation of mobile manipulators. 
\methodname{} unifies multiple human interfaces including RGB and depth cameras, virtual reality controllers, keyboard, joysticks, etc., and any combination thereof.
In its more accessible version, \methodname{} works using simply vision (e.g., an RGB-D camera), lowering the entry bar for humans to provide mobile manipulation demonstrations.
We demonstrate the versatility of \methodname{} by teleoperating several existing mobile manipulators --- PAL Tiago++, Toyota HSR, and Fetch --- in simulation and the real world. We demonstrate the quality of the demonstrations collected with \methodname{} by training imitation learning policies for mobile manipulation tasks involving synchronized whole-body motion.
Finally, we also show that \methodname{}'s teleoperation channel enables teleoperation \textit{on site}, looking at the robot, or \textit{remote}, sending commands and observations through a computer network, and perform user studies to evaluate how easy it is for novice users to learn to collect demonstrations with different combinations of human interfaces enabled by our system.
We hope \methodname{} becomes a helpful tool for the community enabling researchers to collect whole-body mobile manipulation demonstrations. For more information and video results, \textcolor{cerulean}{\url{https://robin-lab.cs.utexas.edu/telemoma-web/}}.
\end{abstract}
{\blfootnote{{Correspondance: \url{sdass@utexas.edu}}}}

\IEEEpeerreviewmaketitle
\section{Introduction}
\label{s_intro}

A core goal of robotics is to build generalist robots capable of operating alongside humans in their environment. To this end, learning from human-collected robot demonstrations has shown promise in endowing robots with the capabilities to solve complex tasks~\cite{argall2009survey, ravichandar2020recent, robomimic2021}, boosted recently by the advent of foundation models capable of learning from large amounts of data~\cite{bommasani2021opportunities, firoozi2023foundation}. While these models demonstrate an impressive semantic understanding of the tasks~\cite{team2023octo, brohan2023rt, open_x_embodiment_rt_x_2023, driess2023palm, chebotar2023q}, these successes have been largely limited to stationary manipulation. However, a large fraction of the tasks that we would like generalist robots to perform require a combination of manipulation and mobility: e.g., sweeping the floor requires moving the broom with both hands and walking around to reach the dirty spots; covering a table with a tablecloth requires holding the tablecloth and pulling it over the table while simultaneously moving to reach all edges.

\begin{figure}[t]
    \centering
    \includegraphics[width=1.0\columnwidth]{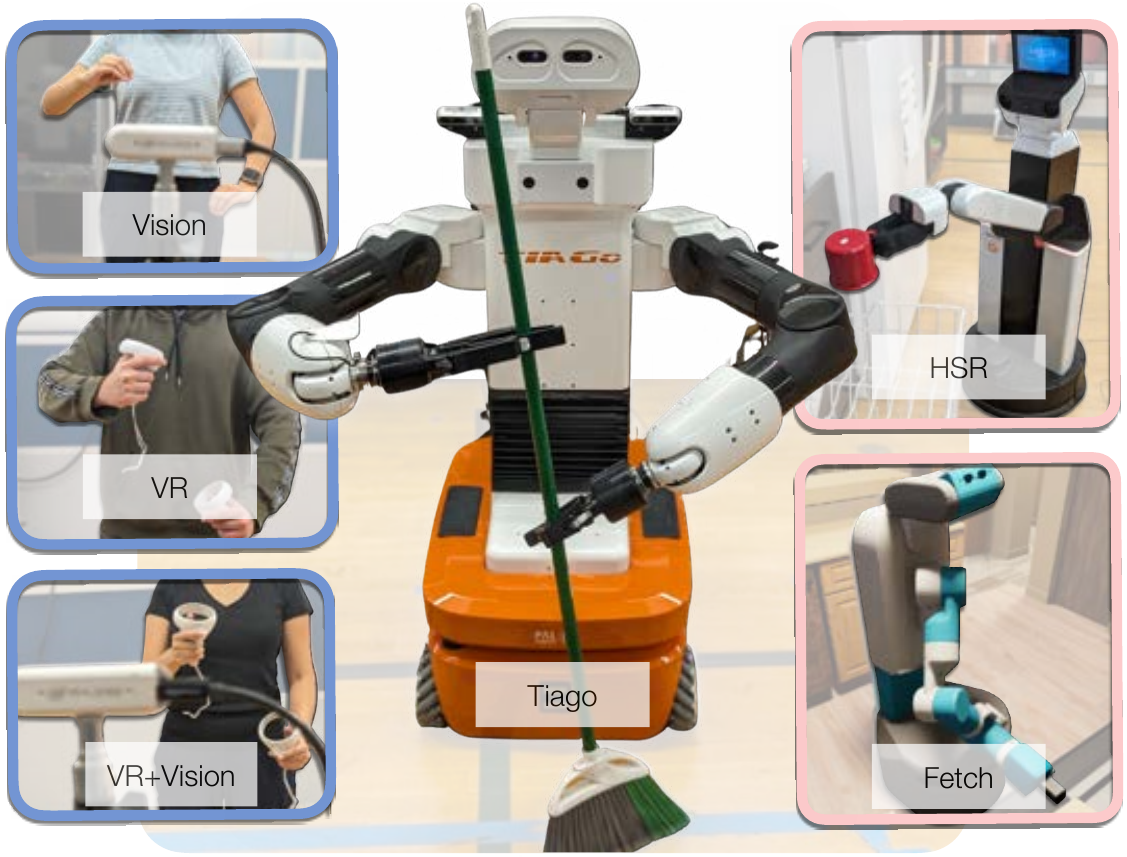}
    \caption{\textbf{\methodname:} a modular and versatile mobile manipulation teleoperation system. \textit{Left and Middle} Demonstrators performing a bimanual sweeping task with the vision-only, virtual reality (VR), and a combination of vision+VR interfaces. \methodname{} enables multiple human interfaces and their combination. \textit{Middle and Right} Tiago (real), HSR (real), and Fetch (simulation), three of the robot platforms that we demonstrate teleoperated for different mobile manipulation tasks with \methodname{}, demonstrating its versatility.}
    \vspace{-1em}
    \label{fig:teaser} 
\end{figure}

One of the reasons why stationary manipulation has enjoyed the benefits of large models, while mobile manipulation has not, is due to the availability of large datasets of human-collected demonstrations~\cite{open_x_embodiment_rt_x_2023, walke2023bridgedata}. They were obtained due to the multiple existing and easy-to-use teleoperation frameworks for stationary manipulators~\cite{mandlekar2018roboturk, zhao2023learning, wu2023gello, qin2023anyteleop, dass2022pato}. For mobile manipulation, however, the existing stationary manipulation teleoperation systems are not sufficient, due to the additional degrees of freedom that the user has to control including mobility and possibly multiple arms.

Several teleoperation frameworks for mobile manipulation have been proposed in the past, with different capabilities and limitations.
They either enable accurate control with specific (and often expensive) hardware like motion capture systems~\cite{arduengo2021human, stanton2012teleoperation, AAMASWS10-setapen} or puppeteering interfaces~\cite{fu2024mobile, purushottam2023dynamic}, or achieve scalability by overloading simple and available devices that work for stationary manipulators such as gamepads~\cite{dass2022pato}, virtual reality controllers~\cite{seo2023deep}, or mobile phones~\cite{tung2021learning,wong2022error}, limiting the expressiveness of the demonstrations. Teleoperation based solely on vision~\cite{zhang2018real, qin2023anyteleop, handa2020dexpilot} promises an available and accessible interface at the cost of accuracy and dexterity. Each device alone presents a tradeoff between accuracy and availability, versatility and expressiveness, and as a result, no single device enables scalable, expressive teleoperation for all mobile manipulators. 

Inspired by the complementary capabilities of several of the human interfaces for teleoperation, we introduce \methodname{} (\textbf{Tele}operation for \textbf{Mo}bile \textbf{Ma}nipulation), a novel teleoperation framework focusing on modularity and versatility. \methodname{} enables users to teleoperate different mobile manipulators with a variety of human interfaces or combinations thereof, in simulation or the real world, providing users the means to select the combination that best fits their teleoperation needs. \methodname{} offers the lowest entry point for researchers: it enables whole-body teleoperation with just a depth camera. But its modularity enables us to use other interfaces such as virtual reality (VR) controllers, a keyboard, a 3D mouse, a mobile phone, or their combination with vision, overcoming the limitations of each individual input modality.
Such a system enables researchers to collect demonstrations of whole-body mobile manipulation tasks at scale for virtually any robot and hardware interface available. We demonstrate that \methodname{} allows researchers to teleoperate different mobile manipulation platforms out-of-box such as PAL Tiago~\cite{pages2016tiago}, Toyota HSR~\cite{yamamoto2019development} and Zebra Fetch~\cite{wise2016fetch}. \methodname{} extends also to simulation, which we demonstrate by integrating it with OmniGibson~\cite{li2023behavior} and the BEHAVIOR-1K benchmark. 

In our experiments, we evaluate both \methodname{}'s usability and its suitability for data collection for imitation learning.
We conducted a user study to evaluate the benefits of modularity in \methodname{} and accessibility of \methodname{}-enabled interfaces for novice users. Our results indicate that a hybrid vision-VR interface is an efficient and natural mode of teleoperation, and that novice users are quickly able to learn to use it. We also successfully trained several imitation learning policies on the data collected using \methodname{} and explored relevant questions for IL for mobile manipulation such as (a)~\textit{What inputs matter in imitation learning for mobile manipulators?}, and (b)~\textit{How do the policies scale as we increase the size of the training data?} We measured the role that different embodiments with different capabilities and the sim-real gap have in teleoperation performance, and demonstrate remote teleoperation of real robots with an analysis of robustness to the latency and delays in the communication. 

In summary, \methodname{} is a novel, modular, and versatile teleoperation framework for whole-body mobile manipulation that facilitates the integration of different human interfaces, robot platforms, and simulators. We hope that our contribution lowers the barrier of entry for researchers to collect demonstrations for imitation learning for mobile manipulation.
\label{s_rw}
\section{Related Work}
\begin{table*}[t]
\caption{Comparison of Existing Mobile Manipulation Teleoperation Systems}
\begin{adjustbox}{width=2\columnwidth,center}
\begin{tabular}{cccccccccc}
\toprule
 &  \multicolumn{3}{c}{\textbf{Teleoperation Support}} & \multicolumn{6}{c}{\textbf{Robot Support}} \\
\cmidrule[0.4pt](lr{0.125em}){2-4}%
\cmidrule[0.4pt](lr{0.125em}){5-10}%
 & Cost / & \multirow{2}{*}{Modular} & \multirow{2}{*}{Modality} & \multirow{2}{*}{Bimanual} & Height  & Whole-Body & Robot & Action & \multirow{2}{*}{Domain}\\
& Accessibility & & &  & Control & Teleop & Agnostic & Space & \\
    \midrule
    \citet{arduengo2021human}  & \faDollar\faDollar & \no & Mocap & \no & \yes & \yes & \yes &  EE Pose / Base Vel. & Real  \\
    \midrule
    
    MoMaRT~\cite{wong2022error}  & \faDollar & \no & Phone & \no & \no & \no & \yes & EE Pose / Base Vel.  & Sim \\
    \midrule
    
    MOMA-Force~\cite{yang2023moma}  & \faDollar\faDollar\faDollar & \no & Kinesthetic & \no & \no & \no & \yes & EE Pose and Wrench & Real \\
    \midrule
    
    SATYRR ~\cite{purushottam2023dynamic}  & \faDollar\faDollar\faDollar & \no & Puppeteer & \yes & \no &  \yes & \no & Joint Pos. / Base Vel. & Real \\
    \midrule
    
    TRILL~\cite{seo2023deep} & \faDollar & \no & VR & \yes & \no & \yes & \yes & EE Poses / Gait & Sim\&Real \\
    \midrule
    
    Mobile ALOHA~\cite{fu2024mobile}  & \faDollar\faDollar\faDollar & \no & Puppeteer &  \yes & \no & \yes & \no & Joint Pos. / Base Vel. & Real  \\
    \midrule
    \textbf{\methodname{}}  & \faDollar & \yes & * & \yes & \yes  & \yes & \yes & EE Poses / Base Vel. / Joint Pos. & Sim\&Real \\
\bottomrule
\end{tabular}
\end{adjustbox}
\label{tab:related}
\end{table*}

\textbf{Teleoperation for General Robotics.} Teleoperation is almost as old as the field of robotics itself~\cite{vertut1986teleoperations}, with early manipulators being controlled in kinematically identical master-slave systems~\cite{siciliano2008springer} similar to the very recent Mobile ALOHA~\cite{fu2024mobile}. More recently, teleoperation has emerged as a critical means of data collection for imitation learning methods~\cite{argall2009survey, ravichandar2020recent}, as the ability to quickly collect large scale robotic data has become paramount for training large capacity behavior models~\cite{mandlekar2019scaling, hoque2023fleet, dass2022pato}.  Many teleoperation modalities have been proposed to address these challenges, including kinesthetic teaching, joysticks, virtual reality, mobile phones, RGB cameras, exoskeletons, and motion capture. 

Each modality has its benefits and shortcomings. Joysticks (e.g. the SpaceMouse) offer intuitive control of a robot's end-effector(s), but fail to enable joint control or navigation~\cite{ryu2010development}.  Virtual reality enables users to perform tasks from the robot's perspective, but is limited by individual tolerance to motion sickness and does not naturally enable simultaneous locomotion and manipulation~\cite{zhang2018deep, whitney2018ros, rahmatizadeh2018vision, delpreto2020helping, gao2019vrkitchen, lipton2017baxter}. Mobile phones offer scalable data collection, but provide a very limited interface, failing to naturally support joint control or base motion~\cite{mandlekar2018roboturk, mandlekar2019scaling}. 
RGB cameras have been explored as an accessible, scalable medium with limited mobility and range of motion~\cite{handa2020dexpilot, qin2023anyteleop, sivakumar2022robotic}. Exoskeletons and master-slave devices enable dexterous control but are typically platform-specific and costly~\cite{fang2023low, zhao2023learning, fu2024mobile, purushottam2023dynamic}, and do not naturally provide a way to coordinate base and arm motion. Motion capture similarly enables high-quality data collection, but is costly and difficult to scale~\cite{arduengo2021human, AAMASWS10-setapen, stanton2012teleoperation}. Kinesthetic teaching was the predominant teleoperation paradigm for imitation learning for many years~\cite{argall2009survey, ravichandar2020recent, kober_reinforcement_2013}, but fails to enable more complicated bimanual or mobile manipulation tasks. Some works explore the combination of different modalities~\cite{wang2012development, fritsche2015first} but fail to be sufficiently general and extensible. Thus, despite the plethora of available options, there remains a need for a teleoperation system capable of adapting to the needs of mobile manipulation in a scalable, accessible way. 

\textbf{Teleoperation for Mobile Manipulation.}
Recent successes in learning from large collections of human demonstrations has been limited to stationary manipulators~\cite{open_x_embodiment_rt_x_2023, team2023octo, brohan2023rt} or simple mobile manipulation tasks like pick and place that do not require coordination between base and arm motion~\cite{brohan2023can, chebotar2023q}. 
This is in part due to the lack of accessible and intuitive ways to collect demonstrations for mobile robots. Recently, some methods have tried to address this using specialized hardware, such as motion capture systems~\cite{arduengo2021human, stanton2012teleoperation, krebs2021kit, AAMASWS10-setapen}, exoskeletons~\cite{fu2024mobile, yang2023moma, matsuura2023development, fang2023low} and more sophisticated human-computer interfaces~\cite{schwarz2023robust, lenz2023bimanual}. On the other hand, several works borrow from successful teleoperation interfaces in stationary manipulation, using interfaces such as VR~\cite{galarza2023virtual, seo2023deep, penco2024mixed, garcia2018robotrix, gan2021threedworld, martinez2020unrealrox, kazhoyan2020learning}, kinesthetic teaching~\cite{yang2023moma}, visual motion tracking~\cite{zhang2018real}, keyboard and mouse~\cite{ratner2015web} and mobile phones~\cite{wong2022error}, by modifying them to enable the control of mobile manipulators. 
Although these interfaces are accessible, they lack the granularity necessary to coordinate all degrees of freedom of a mobile robot for a true mobile manipulation task. 

We summarize the main features of \methodname{} and contrast it with related systems in Table~\ref{tab:related}. The criteria for each category is described further in Appendix A. We compare across two primary dimensions: the teleoperation modalities provided, and the robot capabilities enabled. \methodname{} is the only teleoperation system to provide modularity and enable the flexible combination of multiple input modalities. Moreover, it is the only system capable of full whole-body motion (including torso control) that remains accessible and robot agnostic. 
\begin{figure*}[t]
    \centering
    \includegraphics[width=1.0\linewidth]{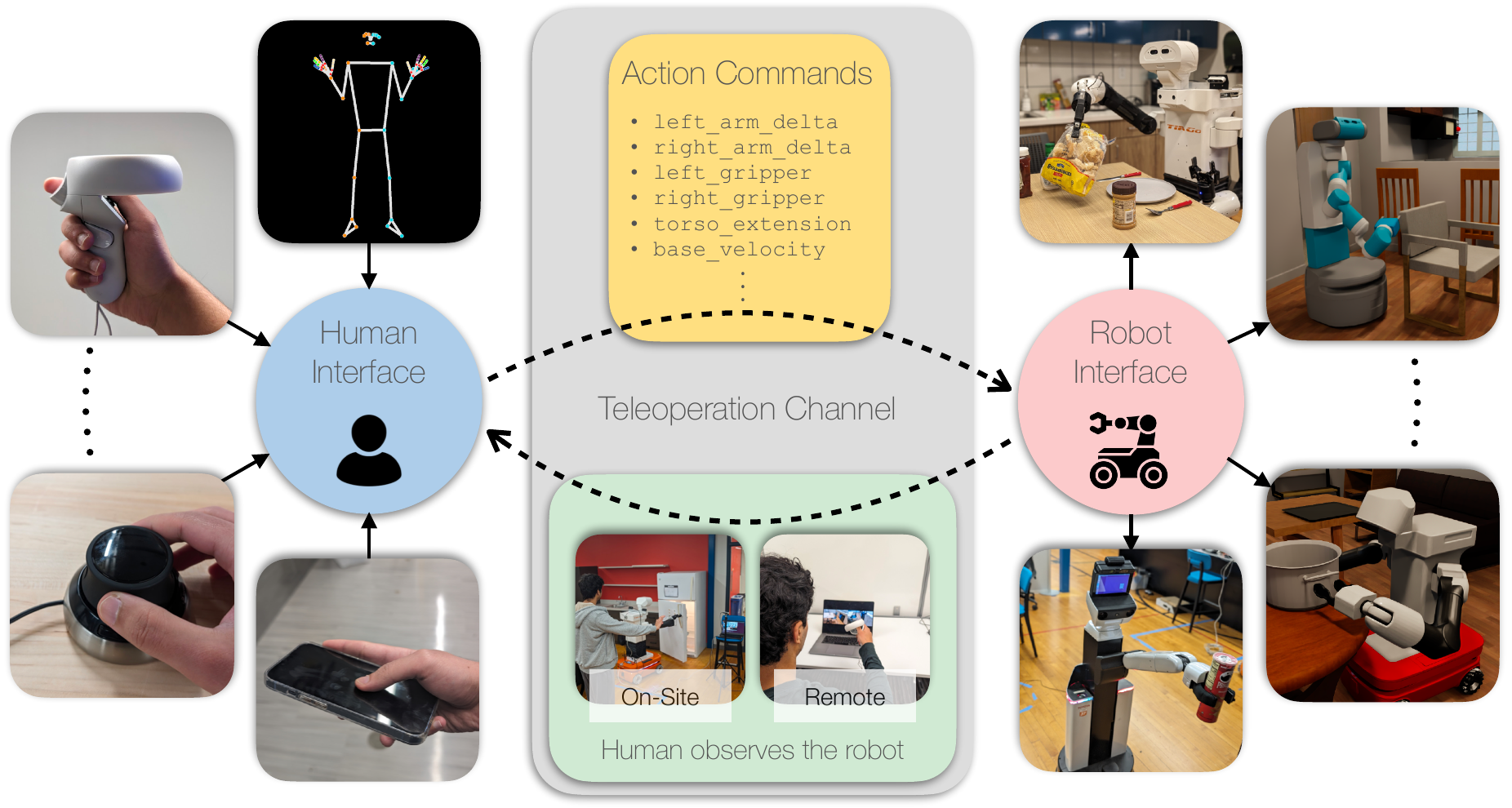}
    \caption{\textbf{\methodname{} System.} \methodname{} consists of three components: the \textit{Human Interface} acquires commands from the human using different input devices; the \textit{Teleoperation Channel} defines the action command structure between the human and the robot interfaces, and, possibly, closes the loop with observations from the robot; and the \textit{Robot Interface} implements a robot-specific mapping of actions to low-level robot commands. This architecture enables modularity and versatility -- combining multiple devices to achieve intuitive whole-body teleoperation for multiple tasks and robots.}
    \label{fig:architecture}
\end{figure*}
\def\taskFigHeight{1.2in}
\def\taskFigWidth{0.44\textwidth}

\begin{figure*}[ht!]
    \captionsetup[subfigure]{aboveskip=-1pt,belowskip=-1pt, justification=centering}
    \centering
    \begin{subfigure}[t]{\taskFigWidth}
        \centering
        \includegraphics[height=\taskFigHeight]{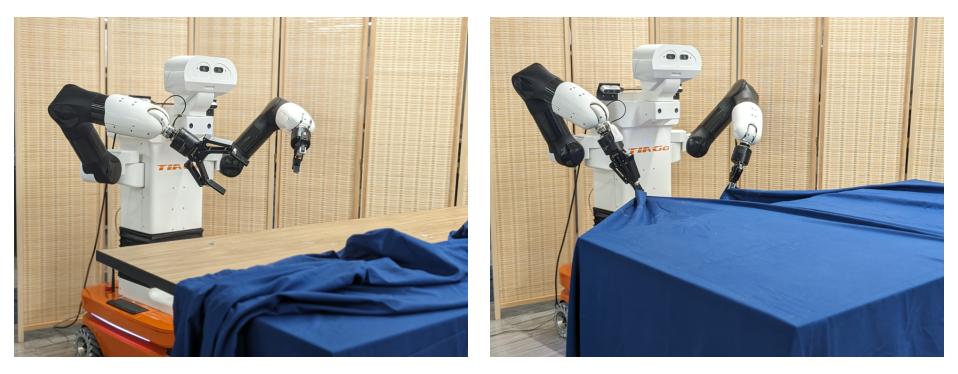}
        \caption{\texttt{Cover table} (Tiago, real world) \\ Grasp the table cloth and drape it over the table}
    \end{subfigure}
    \begin{subfigure}[t]{\taskFigWidth}
        \centering
        \includegraphics[height=\taskFigHeight]{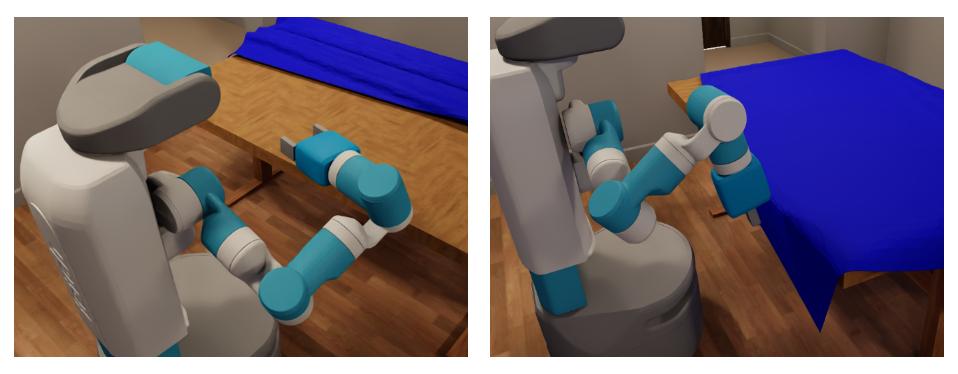}
        \caption{ \texttt{Cover table} (Fetch, simulation)}
    \end{subfigure}
    
    \begin{subfigure}[t]{\taskFigWidth}
        \centering
        \includegraphics[height=\taskFigHeight]{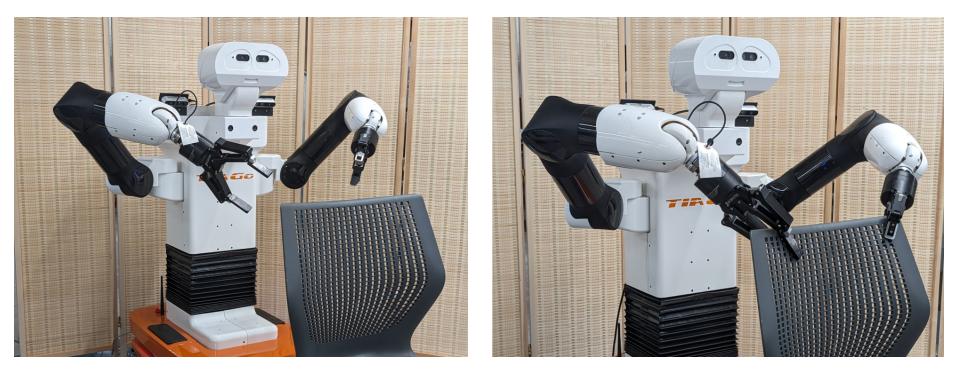}
        \caption{\texttt{Slide chair} (Tiago, real world) \\ Orient towards the chair and push it under the table}
    \end{subfigure}
    \begin{subfigure}[t]{\taskFigWidth}
        \centering
        \includegraphics[height=\taskFigHeight]{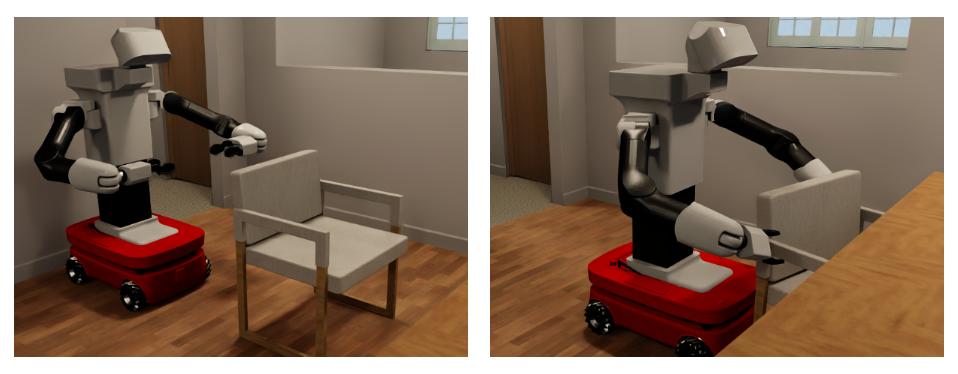}
        \caption{ \texttt{Slide chair} (Tiago, simulation)}
    \end{subfigure}
    
    \begin{subfigure}[t]{\taskFigWidth}
        \centering
        \includegraphics[height=\taskFigHeight]{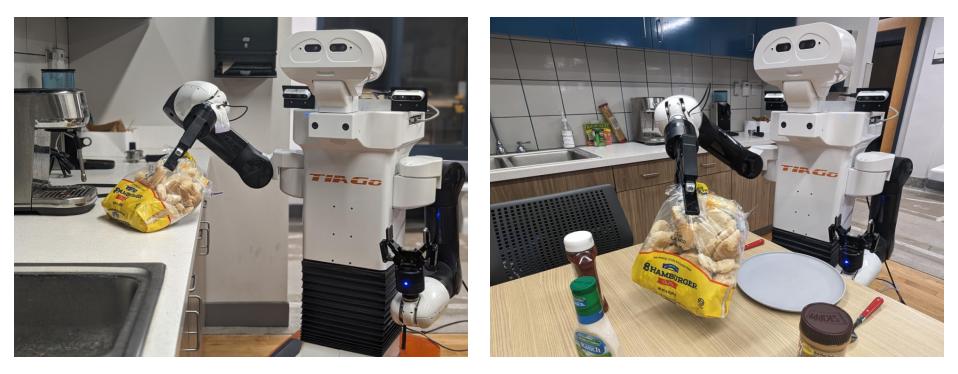}
        \caption{\texttt{Serve bread} (Tiago, real world) \\
        Pick up a packet of bread and deliver to the breakfast table}
    \end{subfigure}
    \begin{subfigure}[t]{\taskFigWidth}
        \centering
        \includegraphics[height=\taskFigHeight]{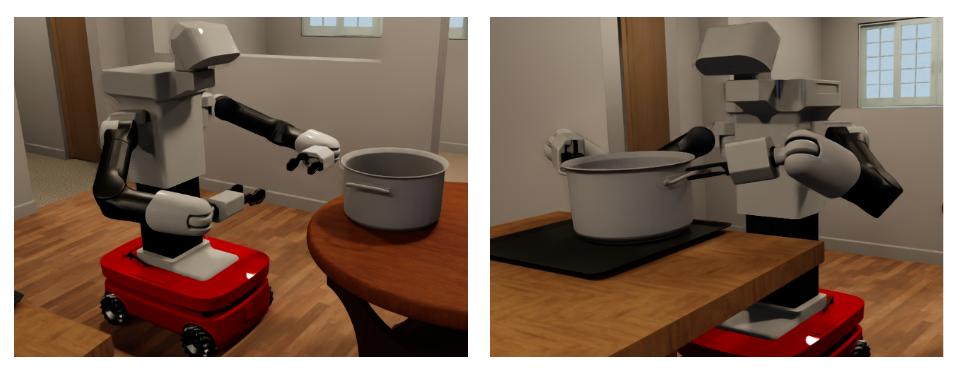}
        \caption{\texttt{Pick pot} (Tiago, simulation) \\ Pick up a pot and transfer to another table}
    \end{subfigure}
    
    \begin{subfigure}[t]{\taskFigWidth}
        \centering
        \includegraphics[height=\taskFigHeight]{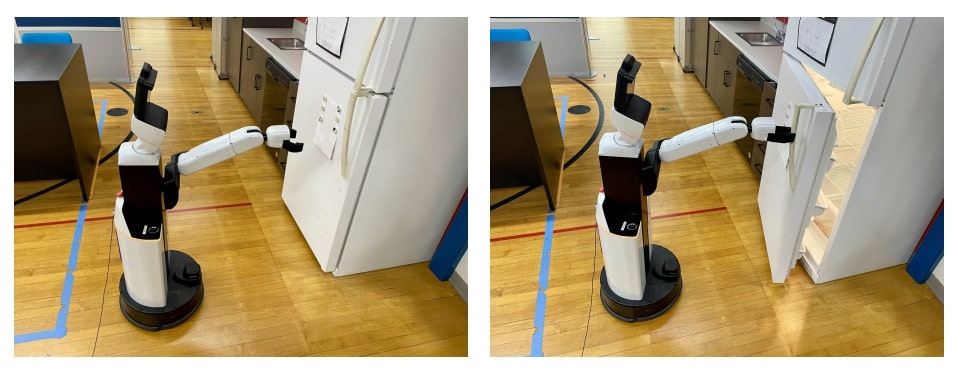}
        \caption{\texttt{Open fridge} (HSR, real world) \\ Open the door of a fridge}
    \end{subfigure}
    \begin{subfigure}[t]{\taskFigWidth}
        \centering
        \includegraphics[height=\taskFigHeight]{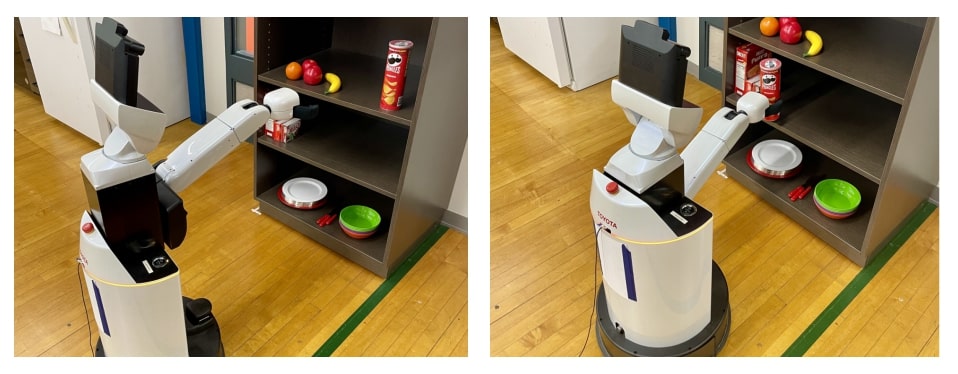}
        \caption{\texttt{Re-shelve chips} (HSR, real world) \\
        Move the misplaced chips to the lower shelf}
    \end{subfigure}
    
    \begin{subfigure}[t]{\taskFigWidth}
        \centering
        \includegraphics[height=\taskFigHeight]{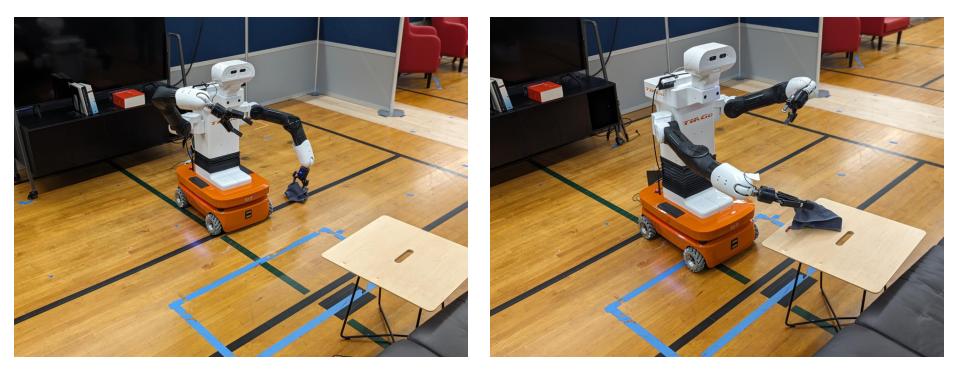}
        \caption{\texttt{Pick up} (Tiago, real world) \\ Grasp a towel from the floor and place it on the table}
    \end{subfigure}
    \begin{subfigure}[t]{\taskFigWidth}
        \centering
        \includegraphics[height=\taskFigHeight]{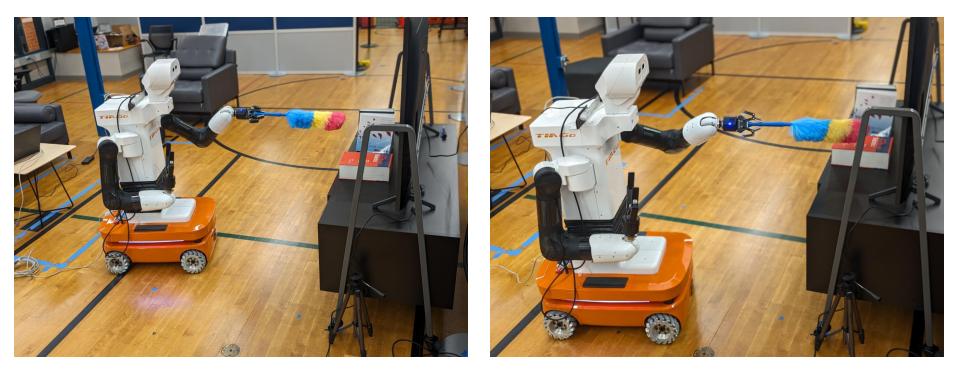}
        \caption{\texttt{Dusting} (Tiago, real world) \\ Dust a
table with books resting on top}
    \end{subfigure}
    
    \caption{\textbf{Tasks in our evaluation of \methodname{}}. Shown above is the initial and goal state of each task.}
    \label{fig:task}
\end{figure*}
\section{\methodname{} System}
\label{s_system}

\methodname{} is a teleoperation system for mobile manipulators --- versatile for different robot morphologies and modular in the human input devices.
It is generally composed of a \textit{Teleoperation Channel} that defines the communication between a \textit{Human Interface} and a \textit{Robot Interface} (Fig.~\ref{fig:architecture}). The \emph{Human Interface} acquires human inputs across different teleoperation modes such as vision, VR, spacemouse, keyboard, and mobile phones, or their combinations, and maps them to a general mobile manipulation action command structure provided by the \emph{Teleoperation Channel} that includes fields such as base, arm, gripper, and torso motion. Multiple input devices can be combined through our \emph{Human Interface} to acquire the action commands in the best suited manner for a task. The \textit{Teleoperation Channel} hands over the action commands to the \emph{Robot Interface}, a robot-specific module that maps the actions to robot motor commands. 
While the specific implementation of the robot interface relies on platform-dependent controllers, our requirements (controllers for the motion of the end-effectors, base, joints, \ldots) are general enough to enable the teleoperation of most existing platforms, in the real world and simulation. 
In the following, we provide additional information about the three components of \methodname{}. 

\subsection{Human Interface}
\label{method:user_interface}
The \emph{Human Interface} is responsible for processing the captured data from various teleoperation input devices and mapping them to a common action command structure. For each input device, the data is processed independently by a device-specific parser that maps the signals from the input modality (keyboard strokes, motion of a VR controller, location of human skeleton keypoints on an image, \ldots) into elements of the teleoperation channel's action command.
\methodname{} supports input modalities such as vision, keyboard, spacemouse, VR~(Oculus Quest and HTC Vive) and mobile phones. In the following, we explain the vision and VR human interfaces from \methodname{} in detail. We test these human interfaces (and their combination) extensively in our experiments because their combined capabilities strike a good balance between availability, generality, dexterity and accuracy. We defer the implementation details of other human interfaces such as spacemouse, keyboard, and mobile phones to Appendix B.

\subsubsection{Vision-Based Human Interface}
\label{method:vision}
\methodname{} offers a unique vision-based pipeline for the whole-body teleoperation of a mobile manipulator using a single RGB-D camera.
We use MediaPipe~\cite{lugaresi2019mediapipe}, a lightweight RGB-based model that executes in real-time for body pose and hand keypoint detection. Our proposed human interface uses the position and rotation of the hips to control the movement of the base of the mobile manipulator. Since the model only provides the relative depth of the keypoints to the center of the hip and not the absolute depth, we use the depth channel of an RGB-D camera to obtain the absolute values.
The hand keypoints are mapped to the end-effector of the robot based on the position and orientation of the palm with respect to the hip.
We compute the per-frame relative pose displacement in Cartesian space of the hands and send them in the teleoperation channel's action command as arm delta commands. Additionally, we use the distance between the center of the hips and ankles to command the robot height for robots with an actuated torso.

\subsubsection{Virtual Reality Controllers as Human Interface}
\label{method:vr}
\methodname{} supports Oculus Quest and HTC Vive virtual reality hardware devices as inputs to the VR human interface. The controllers are tracked with respect to the headset for Oculus and with respect to the lighthouse for HTC Vive. Similar to \citet{seo2023deep}, the tracked hand poses in Cartesian space are used to command the end-effector in the task-space. As in the vision-based interface, we compute the per-frame relative pose displacement of hands and use them in the teleoperation channel’s action command. The joysticks integrated in the VR are used to command the velocities of the mobile base and also control the torso extension.

\subsection{Teleoperation Channel}
The \emph{Teleoperation Channel} defines how the \emph{Human Interface} communicates with the \emph{Robot Interface}, and is the key to \methodname{}'s generality and modularity. 
Specifically, the \emph{Teleoperation Channel} defines an action command structure that serves as a bridge between the human and the robot and the way the active human interfaces populate the entries of this structure.

During deployment, users can specify what input modality they want to use to control each part of the robot's embodiment including left and right arms and hands, torso, and base. The \emph{Teleoperation Channel} automatically manages the action assignment based on the user specification, and  consolidates the possible missing elements of the action commands due to differences in hardware frequency or network delays.

Finally, the \emph{Teleoperation Channel} also defines the mechanism by which humans \textit{close the loop} with the robot and observe the execution of the action commands, adapting those to achieve the mobile manipulation tasks. We consider two methods of observation: on-site and remote. When on-site, the human directly observes the robot executing the action commands. When remote, the \emph{Teleoperation Channel} communicates the images from the onboard sensors of the robot to the human interface to be displayed for the human, enabling teleoperation from a different location. We evaluate both modes in our experiments (Sec.~\ref{s_exp}).

\subsection{Robot Interface}
The \emph{Robot Interface} is a robot-specific module that maps the commands obtained from the \emph{Human Interface} to the motor commands to the robot. In most of our experiments, those are torques at the joints of the robot. The specific controllers used to compute the torques are not part of the \methodname{} system but they are necessary to map the action commands obtained from the \emph{Teleoperation Channel} into low-level commands. We do not deem our requirements for the robot platforms too high: the robot should provide some controllers to move either the end-effector(s) and the base in Cartesian space, the joints, or combinations of both. 

The action command structure in \methodname{} relayed to the \emph{Robot Interface} can either contain values in task-space~(end-effector Cartesian relative motion), joint space~(e.g., torso commands or motion to other joints) and/or velocities~(e.g., base commands), or different combinations of those, as specified by the user during deployment. The \emph{Robot Interface} processes these commands based on the particular robot embodiment, filters out the unusable action components~(such as left hand commands for a single-armed robot like Fetch), and maps the rest to the robot using the preferred choices of controllers such as operational space control~\cite{khatib1987unified} to control one task frame, or whole-body control~\cite{khatib2004whole,mansard2009versatile} to command the entire robot jointly, or having separate controls for each part of the robot.

\section{Experiments}
\label{s_exp}

In our experiments we seek to answer the following questions: (1)~What are the benefits of \methodname{}'s modularity?~(Sec.~\ref{exp:study}) (2)~Can \methodname{} collect high-quality data for imitation learning?~(Sec.~\ref{exp:imitation}), (3)~How does \methodname{} perform in remote teleoperation of the robot with possible network delays?~(Sec.~\ref{exp:remote_teleop}), and (4)~What is the effect of different robot embodiments and the gap between simulation and real in the usability of \methodname{}?~(Sec.~\ref{exp:comparing}).

\subsection{User Study}
\label{exp:study}
To assess the performance of different teleoperation modalities in the \methodname{} framework, we performed two user studies with the PAL Tiago++ robot. We compared three teleoperation modalities described in Sec.~\ref{method:user_interface}: \textit{VR}, in which the user controls the robot's arms with the Oculus controllers and the base and torso with the controller joysticks; \textit{Vision}, in which the user's pose is tracked with an RGB-D camera to control the arms, torso and base motion; and \textit{VR+Vision} combining both modalities, in which the robot's arms are controlled using the Oculus controllers and the base and torso motion is controlled via human pose tracking from RGB-D data. 

In the first user study, we compared the three modalities (\textit{VR}, \textit{Vision}, \textit{VR~+~Vision}) to assess the completion time in two tasks: \texttt{cover table} (Fig.~\ref{fig:task}(a)), in which the robot must grasp a tablecloth with both hands and drape it over a table, and \texttt{dusting} (Fig.~\ref{fig:task}(j)), in which the robot must dust a table with books resting on top. Both tasks, but especially the \texttt{dusting} task, benefit from the simultaneous motion of base and arm(s), i.e., whole-body motion, as enabled by \methodname{} since the robot is required to navigate around the desk while periodically moving the hands to clear out any dust.

\begin{figure}[t!]
    \centering
    \includegraphics[width=1\columnwidth]{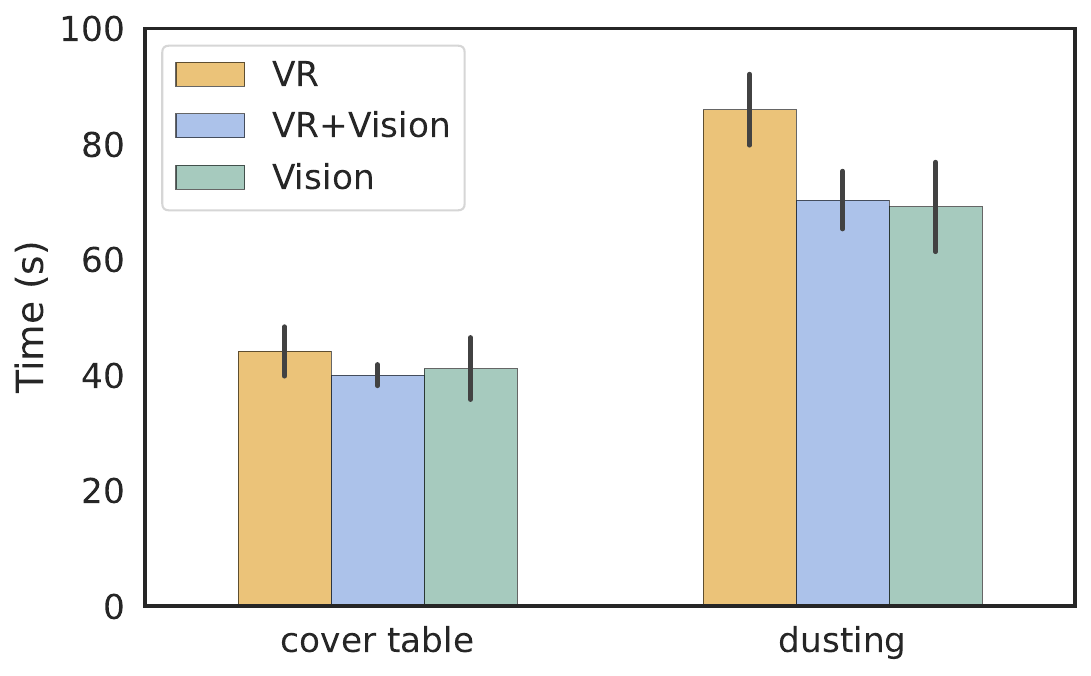}
    \caption{\textbf{User Study 1: Completion Time.} Vision modalities outperform only-VR for the more challenging \texttt{dusting} task. Error bars denote the standard error of the mean.}
    \label{fig:user-time}
\end{figure}

We recruited 12 participants with varying levels of teleoperation experience. Each user was given the same instructions and a brief practice period with each modality. The order in which users received the devices was randomized. The completion times for successful trials are provided in Fig.~\ref{fig:user-time}. The only failures observed occurred with the \textit{Vision} modality (3 fails out of 12 \texttt{dusting} trials) due to noise and inaccuracies in the pose tracking. We observe that in the \texttt{cover table} task, performance is comparable across teleoperation modalities. However, in the \texttt{dusting} task, pure \textit{VR} is generally slower than \textit{VR~+~Vision }or \textit{Vision} alone due to the lack of intuitive whole-body teleoperation: because moving the base requires using the joysticks on the controllers, users tended to only move the arm or the base one at a given time. The results indicate that on their own, both \textit{VR} and \textit{Vision} present drawbacks pertaining to their individual modalities, but when combined in the form of \textit{VR~+~Vision}, \methodname{} can overcome their individual drawbacks to enable an improved teleoperation experience. These results support empirically the importance of enabling multiple input modalities and their combination for teleoperation of mobile manipulators, and \methodname{}'s potential for enabling data collection in more complex mobile manipulation tasks beyond pick and place.  

\begin{figure}[t!]
    \centering
    \includegraphics[width=1\columnwidth]{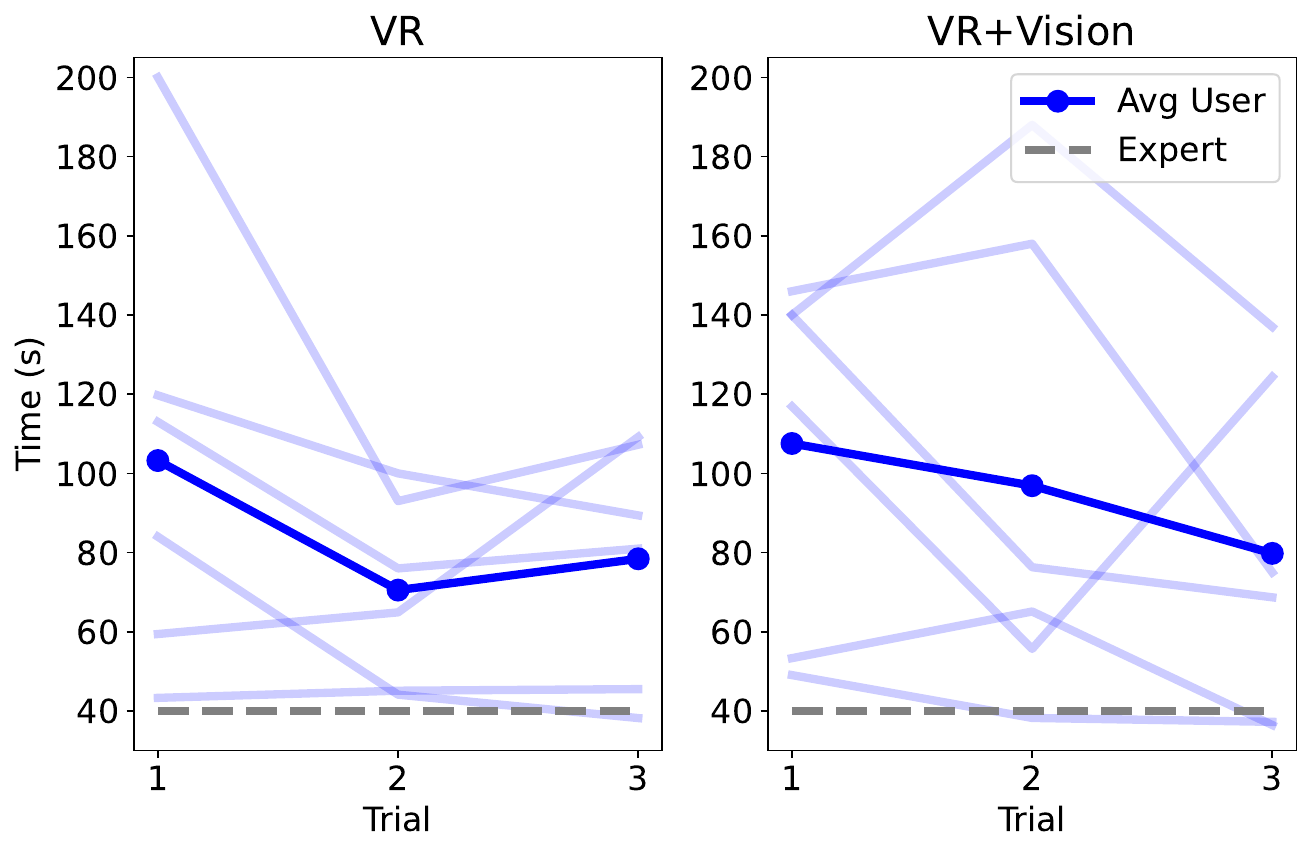}
    \caption{\textbf{User Study 2: User Improvement, Learning Curve.} New users generally improve at completing the \texttt{pick up} task with \methodname{} across teleoperation modalities. Transparent lines show individual learning curves.}
    \label{fig:user-improvement}
    \vspace{-2em}
\end{figure}

In the second study, we sought to assess whether \methodname{} users improve over time by measuring their learning curve. We recruited 6 participants with varying levels of teleoperation experience and compared two modalities (\textit{VR} and \textit{VR~+~Vision}, order randomized) on the \texttt{pick up} task (Fig.~\ref{fig:task}(i)). In this task, the robot must lower its torso in order to grasp a towel from the floor, hand the towel from one hand to the other, navigate to a table, and place the towel on the table. Users completed three consecutive trials with each modality and completion times were recorded. The results are visualized in Fig.~\ref{fig:user-improvement}. We observe that new users generally improve at completing tasks with the system, with an average decrease of 29\% and 26\% in task completion time over three trials for the \textit{VR} and \textit{VR~+~Vision} respectively. The completion times were generally similar between \textit{VR} and \textit{VR~+~Vision}, with some slower times in the \textit{VR~+~Vision} modality owing to the increased difficulty of controlling additional degrees of freedom simultaneously and the additional noise introduced by the vision-based human interface. Despite slowing performance in some trials of this task, the additional capabilities from \textit{VR~+~Vision} have the potential to unlock new whole-body control applications once mastered. Taken together, these two user studies demonstrate the benefits of \methodname{} as a modular teleoperation system. 

\subsection{Imitation Learning with \methodname{}'s Data}
\label{exp:imitation}
To empirically evaluate the quality of the data collected with \methodname{}, we train several visuomotor policies with behavioral cloning~\cite{schaal1999imitation} using the data collected on a Tiago++ robot~(\emph{real}). We consider three diverse mobile manipulation tasks:
\begin{itemize}
    \item \texttt{cover table}: Similar to the one described in Sec.~\ref{exp:study}, the tasks involves bimanual grasping of a tablecloth and draping it over a table~(Fig.~\ref{fig:task}(a)). 
    \item \texttt{slide chair}: A bimanual task, that requires the robot to navigate and align itself behind a chair, grasp it, and push the chair under a table~(Fig.~\ref{fig:task}(c)). 
    \item \texttt{serve bread}: In a real kitchen setting, the robot is required to navigate to the kitchen counter, pick a bag of bread, and deliver it to the breakfast table~(Fig.~\ref{fig:task}(e)). 
\end{itemize}

We collected 50 demonstrations each for \texttt{slide chair} and \texttt{serve bread} tasks and 100 demonstrations for \texttt{cover table} task using the combined \textit{VR~+~Vision} interface of \methodname{}. Additional demonstrations in the \texttt{cover table} were necessary to allow the policies to learn the necessary accurate grasps on the cloth. 

\textbf{Policy Architecture, Observations and Actions.} We used a feed-forward MLP~(BC) and a recurrent LSTM based network~(BC-RNN)~\cite{robomimic2021} with a sequence length of 10. The inputs to all policies included RGB-D images obtained from two realsense cameras attached on each shoulder of the robot, end-effector poses of the hands, gripper state, and the change in the mobile base pose obtained from the odometry of the robot. The policies output a 17-dimensional action space: 6D Cartesian deltas and a gripper command for each of the hands, and linear and angular velocities for the base.

\textbf{Comparing Input Modalities.} To analyze the importance of depth sensing in learning mobile manipulation tasks, we train two sets of policies: the first set was trained exclusively on RGB observations, while the second combined RGB and Depth. The performance of the two sets of policies for each of the tasks is summarized in Table~\ref{tab:il_modality}. Our analysis reveals a consistent trend: irrespective of the policy architecture, the inclusion of depth information markedly enhances performance across all tasks. Qualitatively, we observe that policies trained using depth can position the base better, significantly improving the efficacy of subsequent arm actions. These findings suggest that depth information is a crucial component for the development of effective mobile manipulation policies, and that the strong dependency between base and arm actions is one of the main challenges in IL for mobile manipulation.

\textbf{Performance with Different Amounts of Data.} To investigate how data volume influences policy performance, we experimented with two distinct policy groups: the first group was trained using the complete dataset we gathered for each task, while the second group utilized only 50\% of these collected demonstrations. The results are summarized in Table~\ref{tab:il_data_scale}; we observe that policies trained with the full dataset consistently outperform those trained with half the data, demonstrating the importance of dataset size in imitation learning, especially in this low-data regime. We additionally notice that BC-RNN strictly outperforms regular BC in all tasks, demonstrating the significance of temporal dependencies for learning mobile manipulation tasks. 

In general, the above experiments provide compelling evidence that IL policies trained with data collected using \methodname{} can reliably perform complex mobile manipulation tasks, thus indicating that \methodname{} can facilitate high-quality data collection for imitation learning. We demonstrate more imitation results in the sim environment in Appendix C.

\begin{table}[t]
\vspace{0.2cm}
\centering
\caption{Performance between IL policies trained with RGB vs. RGBD images as inputs. Successes measured over 10 rollouts.}
\begin{tabular}{ccccccc}
\toprule
 &  \multicolumn{2}{c}{\textbf{Cover Table}} & \multicolumn{2}{c}{\textbf{Slide Chair}} & \multicolumn{2}{c}{\textbf{Serve Bread}}\\
\cmidrule[0.4pt](lr{0.125em}){2-3}%
\cmidrule[0.4pt](lr{0.125em}){4-5}%
\cmidrule[0.4pt](lr{0.125em}){6-7}%
Modality & RGB & RGB-D & RGB & RGB-D & RGB & RGB-D\\
\midrule
BC & 60 & 60 & 40 & 60 & 20 & 40\\
BC-RNN & 70 & \textbf{90} & 50 & \textbf{80} & 30 & \textbf{70}\\
\bottomrule
\end{tabular}
\label{tab:il_modality}
\end{table}

\begin{table}[t]
\vspace{0.2cm}
\centering
\caption{IL Policy performance scale with data. Successes measured over 10 rollouts.}
\begin{tabular}{ccccccc}
\toprule
 &  \multicolumn{2}{c}{\textbf{Cover Table}} & \multicolumn{2}{c}{\textbf{Slide Chair}} & \multicolumn{2}{c}{\textbf{Serve Bread}}\\
\cmidrule[0.4pt](lr{0.125em}){2-3}%
\cmidrule[0.4pt](lr{0.125em}){4-5}%
\cmidrule[0.4pt](lr{0.125em}){6-7}%
Fraction of data & 50\% & 100\% & 50\% & 100\% & 50\% & 100\%\\
\midrule
BC & 60 & 60 & 40 & 60 & 30 & 40\\
BC-RNN & 60 & \textbf{90} & 70 & \textbf{80} & 40 & \textbf{70}\\
\bottomrule
\end{tabular}
\label{tab:il_data_scale}
\end{table}
\def\timeFigHeight{2in}

\begin{figure*}[t!]
    \captionsetup[subfigure]{aboveskip=2pt,belowskip=-1pt, justification=centering}
    \centering
    \begin{subfigure}[t]{0.39\textwidth}
        \centering
        \includegraphics[height=\timeFigHeight]{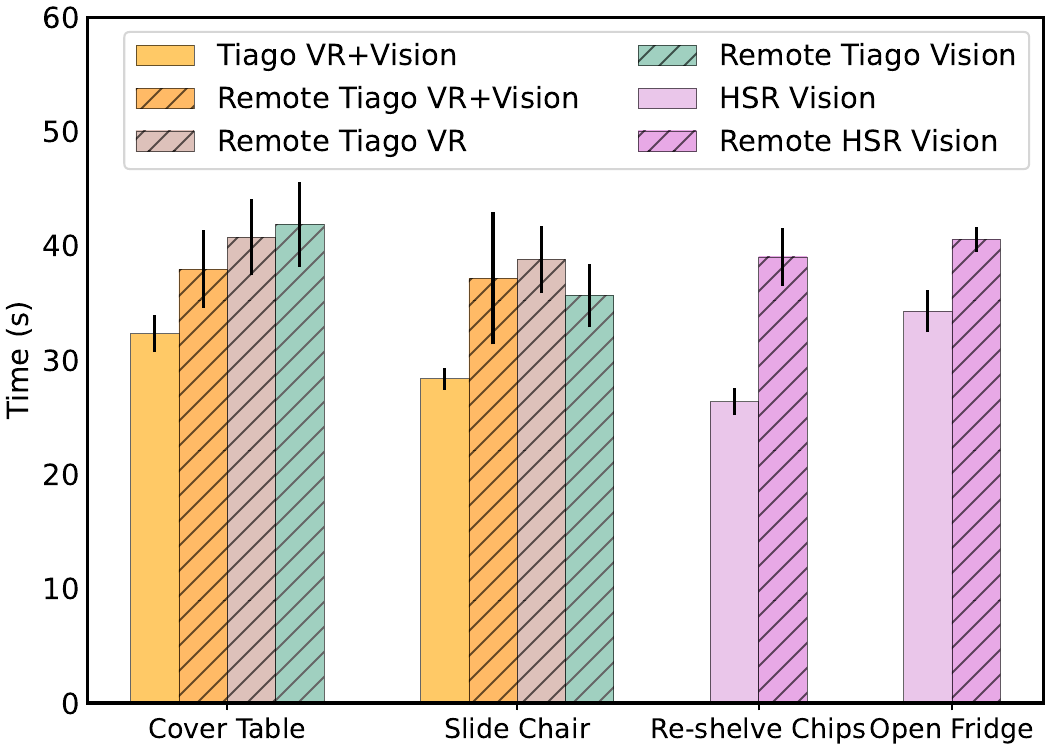}
        \caption{On-Site vs. Remote Teleoperation
}
    \end{subfigure}
    \begin{subfigure}[t]{0.21\textwidth}
        \centering
        \includegraphics[height=\timeFigHeight]{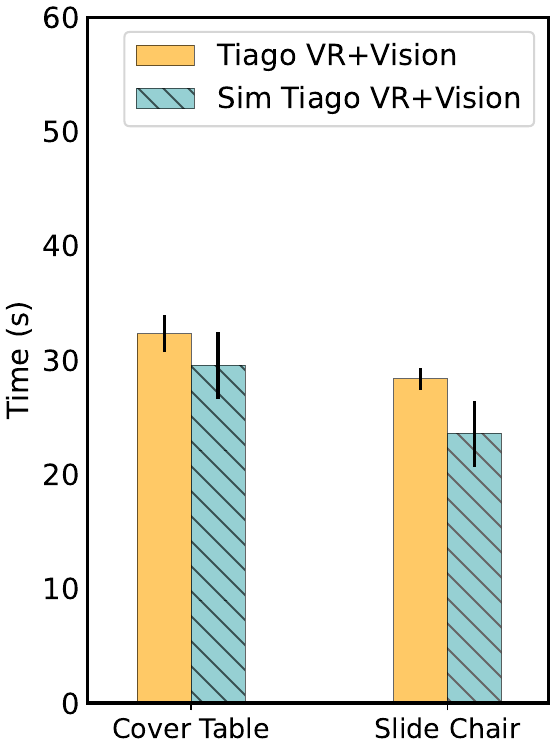}
        \caption{Real world vs. Simulation}
    \end{subfigure}    
    \begin{subfigure}[t]{0.38\textwidth}
        \centering
        \includegraphics[height=\timeFigHeight]{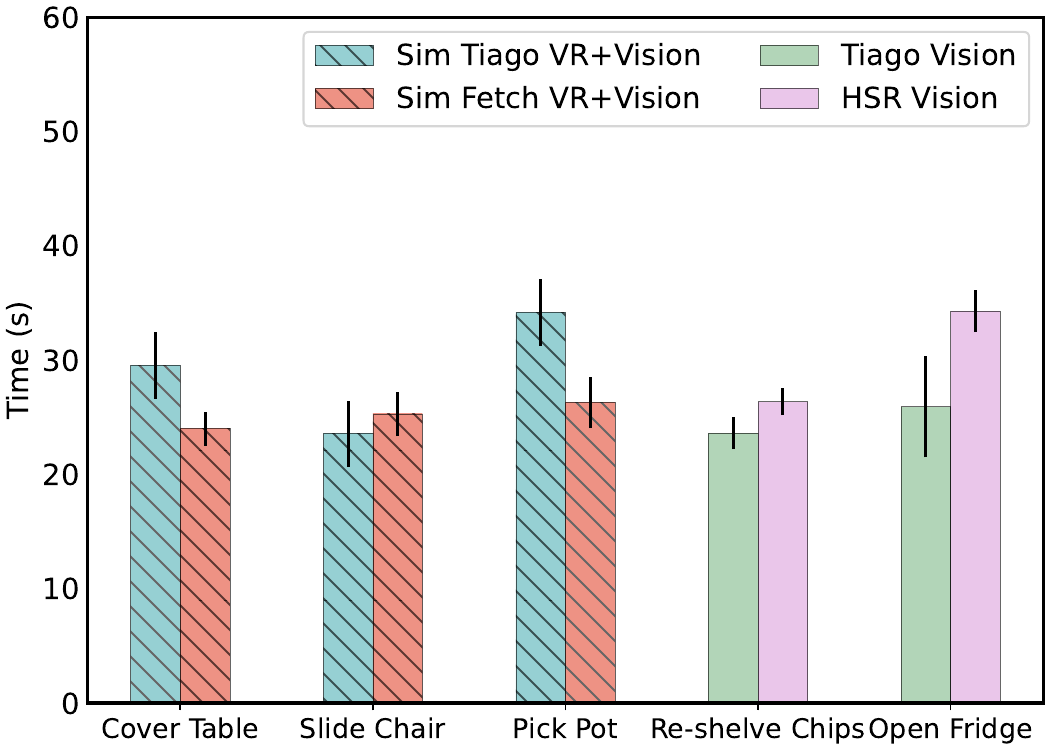}
        \caption{Teleoperating different robot embodiments}
    \end{subfigure}
    
    \caption{Completion times in different experiments with \methodname{}. The bars indicate the mean and standard deviation of several trials (see text). \textit{From left to right:} Comparing completion times for tasks performed on-site and remote, with HSR and Tiago; Completion times for real vs. simulated tasks with Tiago; Completion times for different robot embodiments on the same tasks in the real world and simulation. \methodname{} allows for multiple tasks in simulation and the real world, with several embodiments}
    \label{fig:time_comparison}
\end{figure*}

\subsection{Remote Teleoperation}
\label{exp:remote_teleop}
\methodname{}'s architecture allows a remote demonstrator to control the robot from a client computer connected over the internet. Instead of watching the robot on-site, the demonstrator is provided with camera streams transmitted by the teleoperation channel from the robot's onboard sensors. To minimize communication delays, \methodname{} 1) sends compressed sensor images from the robot and decompresses them on the client, and 2) in the case of a vision-based human interface, \methodname{} processes the RGB-D images from the vision interface on the client side and only sends the action commands over the teleoperation channel. For other interfaces, the demonstrated action commands are directly sent to the \methodname{}'s robot interface.

We demonstrate the remote teleoperation capability of \methodname{} on several combinations of robot hardware and user interfaces. To evaluate the effects of communication delays, we compare the task completion time between on-site and remote demonstrations using Tiago++ and Toyota HSR each on two different tasks. The \texttt{cover table} and the \texttt{slide chair} tasks are completed using Tiago++ with the on-site \textit{VR~+~Vision} interface and three remote interfaces (\textit{VR}, \textit{Vision}, \textit{VR~+~Vision}). The \texttt{re-shelve chips} task, in which the robot must move the misplaced chips to the lower shelf (Fig. \ref{fig:task}(h)), and the \texttt{open fridge} task, in which the robot must open a fridge (Fig. \ref{fig:task}(g)), are completed using HSR with the \textit{Vision} interface. The demonstrations are provided by an expert user of each robot. The Wi-Fi speed is about 100 Mbps as measured on the HSR. Fig.~\ref{fig:time_comparison}(a) shows the completion time in each modality averaged over 3 runs. We observe that remote human demonstrators have slower reaction times due to delays and limited resolutions of the camera streams, but \methodname{} provides the capability to successfully complete the tasks under regular network conditions. We expect to be able to demonstrate these remote teleoperation capabilities from the RSS venue to our lab.

\subsection{Comparing Different Embodiments and Sim vs. Real}
\label{exp:comparing}
In the final set of experiments, we seek to study how the domain~(sim vs. real) and the type of robot~(Tiago vs. HSR and Tiago-sim vs. Fetch-sim) influence the teleoperation behavior for the same tasks.

\subsubsection{Sim vs. Real} Fig.~\ref{fig:time_comparison}(b) depicts the results of comparing completion time for \texttt{cover table} and \texttt{slide chair} tasks in simulation and real environment using a Tiago robot. We use sim time for simulation evaluation because of OmniGibson's sub-realtime soft-body simulation. By maintaining consistency across the robot, the task, and the teleoperation interface, we find that for both tasks the completion time in simulation and real are close, demonstrating that the simulation environment in OmniGibson is a good proxy for mobile manipulation in the real world, and that teleoperating with \methodname{} provides a natural mechanism to collect demonstrations in sim.

\subsubsection{Comparing Embodiments}
We additionally compare how the completion time varies as we change the robot being teleoperated by maintaining the task, teleoperation interface and reality to be consistent. We compare Tiago and HSR on \texttt{re-shelve chips} and \texttt{open fridge} tasks and depict the results in Fig.~\ref{fig:time_comparison}(c, right). We observe that the higher number of degrees of freedom offered by Tiago compared to HSR allows more fluid motion during teleoperation and enables a more efficient (faster) completion of the task.

In simulation, we compare Tiago and Fetch on \texttt{cover table}, \texttt{slide chair}, and \texttt{pick pot} tasks and depict the results in Fig.~\ref{fig:time_comparison}(c, left). For the \texttt{pick pot} task, we enabled sticky grasping (creating a controllable constraint between hand and object) since the task would be infeasible otherwise for a single-armed robot like Fetch. We observe that Fetch is faster than Tiago on tasks requiring table-top manipulations, possibly due to Fetch's larger size and longer arms, making manipulation easier for users.

\section{Conclusions}
\label{s_conc}

We presented \methodname{}, a novel teleoperation system for mobile manipulators that enables versatility through modularity. While no single teleoperation interface provides all benefits of enabling dexterous, whole-body mobile teleoperation while remaining low cost and scalable, our general, modular teleoperation interface provides the ability to combine multiple existing modalities combining also some of the benefits of them. This results in a performant data collection system that scales to many different robots and tasks, as indicated by our user studies, imitation learning, remote teleoperation, and comparisons between embodiments and sim vs. real. We note some limitations of \methodname{}. 
First, when tracking human pose from RGB data, noise and inaccuracies can impact a user's ability to accomplish tasks. Combining vision with a more accurate interface like VR enables accurate arm control and synchronization of base and arm movement, but would still benefit from better visual pose-tracking models. Second, occlusion presents a challenge for the vision-based modalities, as the camera placement has an impact on the operator's visibility of the robot's workspace. This can be mitigated by carefully choosing a camera placement, using multiple cameras, or rendering robot observations on a screen. Extending \methodname{} to incorporate a puppeteering human interface would enable even more accurate tasks at the cost of mobility.

In closing, we have demonstrated \methodname{}, a general, modular, accessible teleoperation system that enables collection of high-quality expert demonstration data for a variety of complex and novel mobile manipulation tasks. We showed \methodname{}'s generality by teleoperating multiple different robots in simulation and reality, and conducted user studies to verify the usability of the system's various modalities. We hope that our system lowers the barrier of entry for researchers to collect high-quality demonstrations for mobile manipulation, and helps unlock new mobile manipulation capabilities.

\bibliographystyle{plainnat}
\bibliography{references}

\clearpage
\newpage
\appendix

\subsection{Criterion for Table~1}
We provide a detailed explanation for each column of Table~1, including the criteria used to categorize methods.
\begin{itemize}
    \item Teleoperation Support
    \begin{enumerate}
        \item \emph{Cost / Accessibility:} We identified three tiers of price based on commercially available systems or disclosed cost. \\ \\
        \begin{tabular}{cl}
         \faDollar: & \$0 -- 1,000 (VR, Vision, Phone)\\ 
         \faDollar\faDollar: & \$1,000 -- 10,000 (Mocap Systems) \\ 
         \faDollar\faDollar\faDollar: & \$10,000+ (Custom Hardware)\\ \\
        
\end{tabular}

        \item \emph{Modular:} True if the method is modular in the sense that it supports multiple input modalities or combinations thereof. \methodname{} is the only method that meets this criteria.  
        \item \emph{Modality:} Modality describes the human interface used for teleoperation (e.g. virtual reality (VR), puppeteering with a kinematically similar device, motion capture systems (Mocap), etc.). 
    \end{enumerate}
    \item Robot Support
    \begin{enumerate}
        \item \emph{Bimanual:} True if the paper demonstrates bimanual teleoperation. 
        \item \emph{Height Control:} True if the paper demonstrates control of the robot's torso joint. 
        \item \emph{Whole-Body Teleoperation:} True if simultaneous arm and base motion is enabled by the method. 
        \item \emph{Robot Agnostic:} True if the method works for many different robots; false if it is specific to a particular platform. 
        \item \emph{Action Space:} ``EE Pose(s)" denotes control of the robot's end-effector(s) in Cartesian space, whereas ``Joint Pos." indicates joint-space control for the arms and/or torso. Base Vel. indicates control of the base velocity; TRILL \cite{seo2023deep} allows users to select among predefined gaits with a VR controller, denoted ``Gait". MOMA-Force enables teleoperation of end-effector Cartesian pose through kinesthetic teaching and additionally records desired end-effector wrenches, denoted ``EE Pose and Wrench". \methodname{} allows users to control end-effector Cartesian pose, base velocity, and torso joint position; it is also readily extensible to joint control when tracking human pose, but this is left for future work.
    \end{enumerate}
\end{itemize}

\subsection{Method Details}
Following we describe how \methodname{} facilitates the use of mobile phones, spacemouse, and keyboards as part of its \emph{Human Interface}~(Sec.~IV-A). We are also open-sourcing the code to the community to facilitate plug-and-play teleoperation for mobile manipulators to improve data collection efficiency.

\begin{figure}[t!]
    \centering
    \includegraphics[width=1\columnwidth]{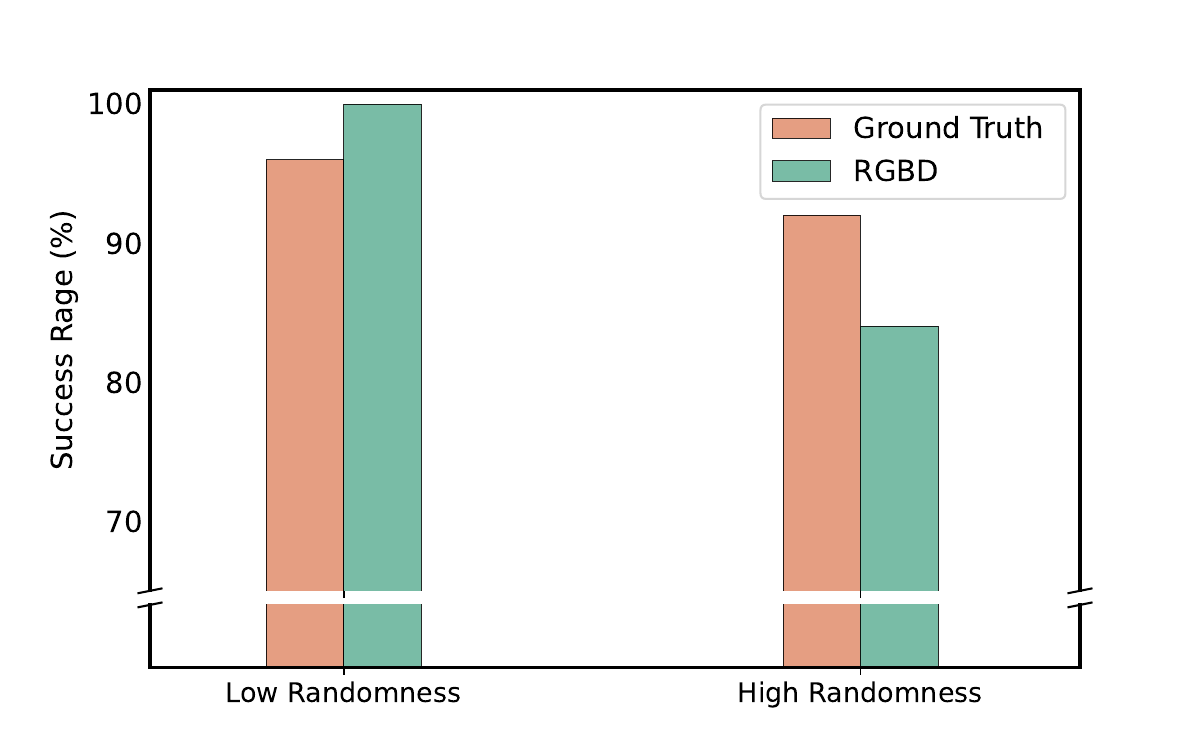}
    \caption{\textbf{IL Results in Simulation.} Policy with RGBD input yields comparable performance to policy with ground truth chair positions as input.}
    \label{fig:il_sim}
\end{figure}

\subsubsection{Mobile Phone} We created an app using the ARKit development kit to track the position and orientation of the mobile phone, which sends commands over the network. Similar to \emph{Virtual Reality Controller}~(Sec.~IV-A2), the end effector is commanded in the task space and the relative pose displacement per frame of the mobile phone is calculated and mapped to the robot end effector. The gripper is controlled by dedicated buttons in the mobile app. Additionally, simultaneous control of left and right arms can be facilitated if two mobile phones are running the app, each phone controlling one of the arms. Mobile phones currently don't support navigation capabilities, but can be combined with other modalities such as the \emph{Vision-based Human Interface}~(Sec.~IV-A1) to facilitate mobile base movements.

\subsubsection{Spacemouse}
Spacemouse has only 6-degrees of freedom, which is why we use mode switching, and control each part of the robot independently. The users can switch modes by pressing one of the side buttons of the spacemouse and switch between controlling left arm, right arm, base and torso. Two spacemouse' can also be used simultaneously for controlling each of the arms and minimizing the mode switching. The displacement of the spacemouse in each of the 6 degrees of freedom is tracked and sent as the delta commands to control the arms. For the base and torso, only the required displacements are used to send commands, while the remaining ones are discarded. The gripper can be toggled by pressing the remaining side button when the spacemouse mode is controlling the corresponding arm. Spacemouse gains significantly from modularity offered by \methodname{}, by minimizing mode switching thus gaining more fluid control of the robot.  

\subsubsection{Keyboard}
Keyboard presses are asynchronously read by the device listeners and each key is mapped to a single DoF of the mobile manipulator. Each key increases / decreases one of the DoFs in the Cartesian space by some preset amount. This results in a large number of keys that the teleoperator has to remember for controlling the robot. Instead, using a smaller set of keys for controlling for instance, just the base, while controlling arms with something more intuitive such as the spacemouse can drastically improve the teleoperation experience on both the interfaces, minimizing the mode switching in case of spacemouse, and reducing the number of keys to keep track of on the keyboard.

\subsection{Imitation Results in Simulation}
\label{appendix:il_sim}
We show the imitation results of the \texttt{slide chair} task in simulation here. We collected 100 demos in OmniGibson, and trained 2 policies using BC with different input observations: one with RGB-D image from the head camera, and the other with oracle chair positions in both world frame and robot base frame from the simulation environment. Robot proprioception, including end effector poses for two arms in base frame, and the base position and velocity in world frame, are also provided as observation input. We evaluated the policy on two task configurations: first with low randomness, where the chair position is uniformly sampled within 0.2 meters parallel to the robot, and second with high randomness, where the sampling interval is 1 meters. Each policy is evaluated with 25 rollouts under these conditions.

The results are shown in Fig.~\ref{fig:il_sim}. We observed that, the performance of policies under high randomness is worse than under low randomness, which is expected because of the increased difficulty. We additionally observe that in both low and high randomness settings, policy trained with RGB-D input performs comparable to the one trained with ground truth chair positions, indicating that the policies are able to extract meaningful environment specific details from images and depth. Qualitatively, we observe that the causes of failure include misalignment between the robot and the chair, slippage of robot grippers, and knocking over the chair due to the application of excessive force. 

\begin{table}[t]\centering
\begin{tabular}{|l|c|}
    \hline
    Hyperparameters & Value \\
    \hline
    \multicolumn{2}{|l|}{\textbf{Behavior Cloning (BC)}}\\
    \hline
    train steps (x500) & 500\\
    batch size & 32\\
    optimizer & Adam\\
    learning rate & 1e-4\\
    image \& depth encoder & resnet-18\\
    policy (w x d) & 512x2\\
    action parameterization & GMM \\
    \hline
    
    \multicolumn{2}{|l|}{\textbf{Recurrent BC (BC-RNN)}}\\
    \hline
    train steps (x500) & 500\\
    batch size & 16\\
    optimizer & Adam\\
    learning rate & 1e-4\\
    image \& depth encoder & resnet-18\\
    LSTM hidden dim & 1000 \\
    LSTM num. layers & 2 \\
    skill horizon & 10 \\
    action parameterization & GMM \\
    \hline
\end{tabular}
\caption{Hyperparameters for the imitation policies (the hyperparameter values were kept consistent across tasks)}
\label{tab:hyperparam}
\end{table}
\subsection{Imitation Learning Policy Hyperparameters}
We performed imitation learning on one simulated~(\texttt{slide chair} -- Appendix Sec. C) and three real world tasks -- \texttt{cover table}~(Fig.~3(a)), \texttt{slide chair}~(Fig.~3(b)) and \texttt{serve bread}~(Fig.~3(c)), that require synchronized hand and base motions. The results and their analysis are presented in Sec.~V-B. We used RoboMimic~\cite{robomimic2021} for training the policies. Comprehensive details of the policy architecture and hyperparameters used for training are provided in Table~\ref{tab:hyperparam}. Note that the same hyperparameters were used across all tasks, and across simulation and real environments. 

Furthermore, in an effort to facilitate and encourage ongoing research in mobile manipulation, the dataset collected on all the tasks will be made available along with the code.
\end{document}